\UseRawInputEncoding
\pdfoutput=1
\documentclass[journal]{IEEEtran}
\usepackage{subfigure}
\usepackage{booktabs}
\usepackage{multirow}
\usepackage[normalem]{ulem}
\useunder{\uline}{\ul}{}
\usepackage{amssymb}
\usepackage{amsmath,amsfonts}
\usepackage{algorithm}
\usepackage{algorithmic}
\usepackage{caption}
\usepackage{array}
\usepackage[caption=false,font=normalsize,labelfont=sf,textfont=sf]{subfig}
\usepackage{textcomp}
\usepackage{stfloats}
\usepackage{url}
\usepackage{verbatim}
\usepackage{graphicx}
\usepackage{cite}
\usepackage{pifont}
\usepackage{enumerate}
\begin{document}
\title{Label Structure Preserving Contrastive Embedding for Multi-Label Learning with Missing Labels}

\author{Zhongchen~Ma\IEEEauthorrefmark{2},\thanks{$\dagger$ Both authors contributed equally to this research.} Lisha~Li\IEEEauthorrefmark{2}, Qirong~Mao,~\IEEEmembership{Senior Member, IEEE,}
and Songcan~Chen\IEEEauthorrefmark{1},~\IEEEmembership{Senior Member, IEEE} 
\thanks{* Songcan~Chen is the corresponding author.}

\thanks{Zhongchen Ma, Lisha Li, Qirong Mao are both with The School of Computer Science and communications Engineering, Jiangsu University, Zhenjiang 212013, China and Jiangsu Engineering Research Center of Big Data Ubiquitous Perception and Intelligent Agriculture Applications, China.
(e-mail: zhongchen\_ma@ujs.edu.cn, lsha.li@outlook.com, mao\_qr@ujs.edu.cn).}
\thanks{Songcan Chen is both with MIIT Key Laboratory of Pattern Analysis and Machine Intelligence, China
and College of Computer Science and Technology, Nanjing University of Aeronautics and Astronautics (NUAA), Nanjing 211106, China (e-mail: s.chen@nuaa.edu.cn).}
}

\markboth{IEEE TRANSACTIONS ON IMAGE PROCESSING,~VOL.~xx,~2022}%
{Shell \MakeLowercase{\textit{et al.}}: Label Structure Preserving Contrastive Embedding for Multi-Label Learning with Missing Labels}

\maketitle

\begin{abstract}
Contrastive learning (CL) has shown impressive advances in image representation learning in whichever supervised multi-class classification or unsupervised learning. However, these CL methods fail to be directly adapted to multi-label image classification due to the difficulty in defining the positive and negative instances to contrast a given anchor image in multi-label scenario, let the label missing one alone, implying that borrowing a commonly-used way from contrastive multi-class learning to define them will incur a lot of false negative instances unfavorable for learning. In this paper, with the introduction of a label correction mechanism to identify missing labels, we first elegantly generate positives and negatives for individual semantic labels of an anchor image, then define a unique contrastive loss for multi-label image classification with missing labels (CLML), the loss is able to accurately bring images close to their true positive images and false negative images, far away from their true negative images. Different from existing multi-label CL losses, CLML also preserves low-rank global and local label dependencies in the latent representation space where such dependencies have been shown to be helpful in dealing with missing labels. To the best of our knowledge, this is the first general multi-label CL loss in the missing-label scenario and thus can seamlessly be paired with those losses of any existing multi-label learning methods just via a single hyperparameter. The proposed strategy has been shown to improve the classification performance of the Resnet101 model by margins of 1.2\%, 1.6\%, and 1.3\% respectively on three standard datasets, MSCOCO, VOC, and NUS-WIDE.  Code is available at \url{https://github.com/chuangua/ContrastiveLossMLML}.
\end{abstract}


\IEEEpeerreviewmaketitle

\section{Introduction}
\IEEEPARstart{M}{ulti-label} learning, handling instances with multiple semantic labels simultaneously, has attracted increasing research interest in various domains such as medical image diagnosis \cite{Ge18}, bioinformatics \cite{10.1109/TKDE.2006.162}, and video annotation \cite{10.1145/1291233.1291245}. For different fields, researchers have proposed several multi-label classification algorithms, such as, Binary Relevance \cite{10.1007/s11704-017-7031-7} which transforms the multi-label classification problem into one or more single-label classification problems, Ensembles of Classifier Chains \cite{10.1007/s10994-011-5256-5}, Random k-Labelsets \cite{5567103}, ML-kNN \cite{10.1016/j.patcog.2006.12.019}, ML-DT \cite{10.5555/645805.670013}, and Rank-SVM \cite{10.5555/2980539.2980628}.

The majority of the approaches above presume that each training instance's label set is complete. In real-world applications, however, collecting full labels for supervised training is difficult. On the one hand, collecting a detailed label set for each image necessitates a significant amount of manpower and material resources. To save the label cost, \cite{https://doi.org/10.48550/arxiv.1705.05640,https://doi.org/10.48550/arxiv.1805.00932,8237359} automatically generate labels by using network supervision, which inevitably leads to the incompleteness of labels. On the other hand, human annotators may ignore labels that they do not know or are not interested in. Consequently, a weakly supervised learning problem \cite{Zhou2018ABI} arises from such an incomplete label collection for each instance, and the traditional multi-label classification algorithms often perform unsatisfactorily in this situation.

How to learn effective feature representation of samples in the absence of labels is one of the key factors for successful learning of multi-label learning with missing labels (MLML) task. In recent years, more and more attention has been paid to the application of CL in self-supervised representation learning, and state-of-the-art performance has been achieved in unsupervised multi-class deep image learning \cite{https://doi.org/10.48550/arxiv.2002.05709,https://doi.org/10.48550/arxiv.1911.05722,caron:hal-02883765}. The present pairwise contrastive methods are far ahead of the traditional contrastive methods, such as max-margin \cite{https://doi.org/10.48550/arxiv.2112.07368}, n-pairs loss \cite{nipsnpairs}, triplet \cite{7298682}, which are particularly outstanding in extracting effective feature representation. \cite{https://doi.org/10.48550/arxiv.2004.11362} extends the self-supervised contrastive learning method to the full supervised task. In the embedding space, samples belonging to the same category are pulled closer while samples of different categories are pushed further. On ImageNet, the proposed loss function shows extremely strong robustness, enabling the model to achieve the best performance in the year. Nevertheless, these CL methods fail to be directly applied to multi-label learning, because it is hard to define the positive and negative samples for an anchor image via its multiple labels. To leverage CL for better performance in multi-label image classification, \cite{contr} proposed a new multi-label classification framework for contrastive learning in a fully supervised setting, which learns multiple representations of an image under the context of different labels from a label-wise perspective. 

However, using CL to build a deep neural network with effective feature representation in MLML is challenging. On one hand, the multi-label datasets often contain a substantial number of false negative labels due to the influence of label missing, such that borrowing the above supervised multi-label learning methods to define positive instances and negative instances for contrasting a given anchor image will incur a lot of false negative instances unfavorable for learning. For example, suppose two instances $x_1$, $x_2$ have the same ground truth label set $\{A, B\}$, but the annotator only labels $A$ as the label set of $x_2$. In contrastive learning, if we define positive and negative instances of instance $x_1$ separately for each label, then for label $B$, $x_2$ becomes a false negative contrastive instance of $x_1$, and vice versa, thus leading to false contrastive learning. On the other hand, to the best of our knowledge, there have not currently had supervised multi-label contrastive learning methods yet that incorporate label dependencies into the learning of MLML tasks, where such dependencies are known to be not only useful but also crucial to solve the problem of missing labels particularly. In our previous work \cite{MA2021107675}, we discover two useful bits of information when investigating the label dependencies: a) low-rank local label dependency, and b)  high-rank global label dependency. In multi-label learning, low-rank local label dependency can provide a lot of effective information for classification \cite{10.5555/2900728.2900863,10.1016/j.eswa.2013.10.030}. 
For example, if an object is labeled as ``grape", it must also be labeled as ``fruit", but the reverse is not necessarily true. Mathematically, the rank of the label sub-matrix for samples that share the same label is smaller than the rank of the original label matrix consisting of all samples. High-rank global  label dependency is also an important part. For the label matrix containing missing items, if we set the rank of the label matrix to be very small, which means that the diversity of labels is limited, then undoubtedly a lot of useful discrimination information will be lost. Li et al. \cite{https://doi.org/10.48550/arxiv.2005.00976} proved that the high-rankness of the label matrix can effectively provide more discrimination information.


In this paper, we design a label structure preserving contrastive loss for multi-label learning with missing labels (CLML). In order to better adapt to the multi-label with missing labels scenario, CLML first elegantly generates positives and negatives for individual semantic labels of an anchor image with the help of a label correction mechanism to identify missing labels, then defines a unique contrastive loss for multi-label image classification with missing labels, i.e., CLML, the loss makes the anchor image more compact not only with its true positive instances but also with its false negative instances. At the same time, CLML naturally preserves low-rank global and local  label dependencies, allowing the model to fully utilize the label matrix's structural information. We show that combining CLML with the existing loss function via a single hyperparameter consistently enhances the model's classification performance on three benchmark datasets. Our contributions can be summarized as follows:

\begin{itemize}
	\item A unique contrastive loss for multi-label image classification with missing labels (CLML) is proposed, the loss is able to accurately bring images close to their true positive images and false negative images, far away from their true negative images.
	
	\item The global and local label dependencies are naturally preserved in CLML, allowing the label correlation to be employed more effectively to solve the multi-label classification task with missing labels.
	
	\item To our knowledge, CLML is the first loss function created specifically for multi-label learning with missing labels utilizing contrastive learning.
	\item We conduct extensive experiments on three benchmark datasets MSCOCO, VOC, and NUS-WIDE, and the results demonstrate the superiority of the proposed framework. 
\end{itemize}

The remainder of this paper is organized as follows. In Section 2, we briefly review some related works. In Section3, the proposed framework is presented. Experimental results are presented in Section 4. Finally, Section 5 summarizes the works of this paper.

\section{Related Work}
The background of multi-label learning with missing labels is presented in this section, along with some existing work on the MLML tasks. Subsequently, we briefly review the current state of contrastive learning in multi-label learning.

\subsection{Multi-label Learning with Missing Labels}
\begin{figure}[!ht]
\centering
\includegraphics[width=.8\linewidth]{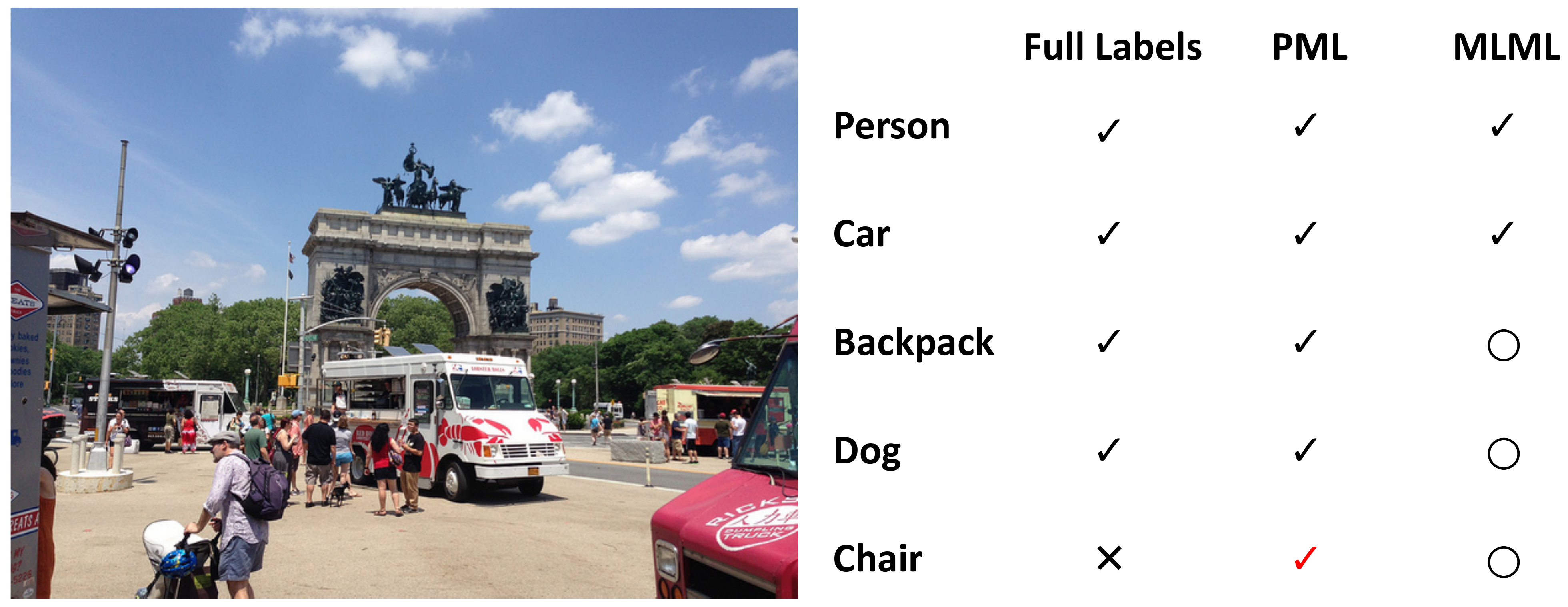}
\caption{$\checkmark$, ×, $\circ$ means a label is present, absent, unknown, and falsely marked label is in red.}
\label{a}
\end{figure}

Compared with traditional learning problems, it is very difficult to label multi-label data, and a larger label space brings higher labeling costs. Therefore, a popular trend in multi-label learning in recent years is to focus on how to build better learning models under weakly supervised information. There are two popular related research topics of weakly supervised machine learning frameworks \cite{Liu_2021}, Partial Multi-Label Learning (PML) \cite{10.5555/1953048.2021049}, and Multi-Label Learning with Missing Labels (MLML) \cite{https://doi.org/10.48550/arxiv.1307.5101}. PML considers a class of hard-to-label problems. For an image, it is relatively easy to identify labels such as ``trees" and ``rivers", but it is more difficult to determine the existence of labels such as ``France" and ``Italy". This ambiguity exists due to some objects being hard to identify, and the human annotators may not be professional enough. However, in many cases, the human annotators can probably guess the correct label range, so the PML chooses to have the human annotators provide all possible labels, with the strong assumption that all labels should be included in the candidate label set. In contrast, complete positive labels are not required in MLML. An unannotated label may be a true negative label or a false negative label, i.e. a missing label. Figure \ref{a} shows the difference between the two in detail.

To better adapt to the problem of MLML, the simple and straightforward approach is to assume all the missing labels are negative \cite{10.1109/CVPR.2011.5995734}. However, this approach is not extensible when the number of categories increases, and it ignores correlations between labels and between instances, which may help in recognition. For an image labeled with ``fish" and ``seaweed", chances are it will also be labeled with ``ocean". Therefore, it is generally believed that the effective way to address MLML tasks is to capitalize on the intrinsic correlations between labels. Most of the methods to describe label correlation use the low-rank structure of label matrix or encode label dependency through graph neural networks. \cite{7023448} proposed an integrated framework that captures the complex correlations among labels using a low-rank structure. \cite{10.5555/2999792.2999869} used a matrix completion theory to complete the instance-label matrix by low-rank regularization of label correlation and instance-instance correlation. \cite{7410830} constructed a unified label dependency network through a mixed graph that incorporates instance-level label similarity and class co-occurrence information. \cite{https://doi.org/10.48550/arxiv.1902.09720} utilized curriculum learning-based strategies and graph neural network proposing a scalable and end-to-end learnable model. However, these methods are either limited to the complexity of model implementation or can only capture global label correlations. 

\subsection{Supervised Contrastive Learning }
In image classification, cross-entropy loss is widely used as the loss function in most deep neural networks. Despite its simplicity and appealing performance, it has a great deficiency in improving the discriminant power of deep neural networks. To bridge this gap, supervised contrastive learning, as a means to obtain highly discriminative features, demonstrates its effectiveness in extracting useful representations in latent space for supervised tasks \cite{https://doi.org/10.48550/arxiv.2002.05709,https://doi.org/10.48550/arxiv.1911.05722}. The core idea of supervised contrastive learning is to minimize the distance between representations of a positive pair and maximize the distance between negative pairs. SupCon \cite{https://doi.org/10.48550/arxiv.2004.11362} considered taking multiple positive samples from the same category of the anchor point. GZSL \cite{https://doi.org/10.48550/arxiv.2103.16173} utilized both class-wise and instance-wise supervision through contrastive embedding. Given the attractiveness and promising results of CL in multi-class classification, it is natural to migrate it to multi-label scenarios to improve the performance of the model. \cite{contr} utilized a deep neural network to learn unique embedding representation for each object in each image, thus, constructing a contrastive learning framework on multiple representations of an image. \cite{Malkinski2020MultiLabelCL} designed a noise contrastive estimation learning algorithm for multi-label scenarios, which combined contrastive training and auxiliary training to solve abstract visual reasoning tasks. In practical application, a large amount of data supervision is not accessible, so people pay more and more attention to the learning task under weak supervision. \cite{yan-etal-2021-weakly-supervised} designed a contrastive loss for the weakly supervised scenario, which assigns a higher weight to the hard negative samples to further encourage diversity. WCL \cite{Zheng2021WeaklySC} is a weakly supervised contrastive learning framework consisting of two projection heads, which combines KNN strategy to expand the positive sample set. It no longer regards a single instance as a single category, thus greatly alleviating the class collision problem and effectively improving the quality of the learned representation. To the best of our knowledge, there are currently no supervised multi-label contrastive learning methods on weakly-supervised multi-label learning task. Therefore, we aim to propose a new contrastive embedding method on a specific weakly-supervised multi-label learning task, i.e., MLML.

\section{Proposed Method}
\subsection{Definition}
In the context of multi-label learning with missing labels setting, we aim to predict the complete label vector of image $i$. Let $\mathbf{X} \in R^{N\times W \times H \times 3}$ denote a batch of images, where $N$ is the batch size, $H$ and $W$ are the height and width of the images. For the $i$-th image  $\mathbf{x}_i \in \mathbf{X}$, we denote $\mathbf{y}_i^o \in \{-1, +1\}^{|C|}$ the observed label vector and $\mathbf{y}_i \in \{-1, +1\}^{|C|}$ the ground truth label vector, where $C$ is the size of label set. If $y_{ij}^o = +1$, it indicates that label $j$ is annotated for image $i$, and vice versa. If $y_{ij} = +1$, it indicates the presence of  label $j$ in image $i$, and vice versa. $\mathbf{Z}=\mathbb{E}\left( \mathbf{X};\theta \right)$ is the deep embedding  of $ N\times D$  dimension of input instance sub-matrix $\mathbf{X}$, where $\theta$ is the parameter of the deep embedding network. The goal is to learn a deep neural network $f(\mathbb{E}(\mathbf{x}),\phi)$ to predict accurately the complete label vector $\mathbf{y}$ of image $\mathbf{x}$, where $f$ is a fully connected layer of the deep neural network as a classifier and $\phi$ is the parameter of the classifier.

\subsection{Label Structure Preserving Contrastive Loss}
\begin{figure*}[htbp]
	\centering
	\includegraphics[width=\textwidth]{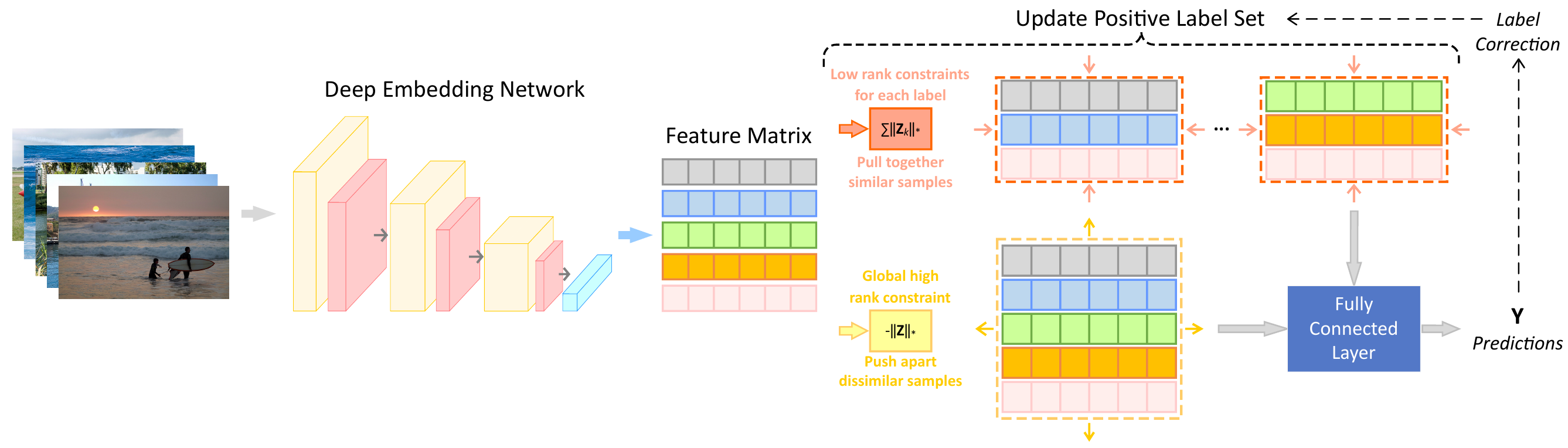}
	\caption{Illustration of proposed contrastive embedding method. We project images into the embedding space through the deep embedding network $\mathbb{E}$ we learned, and then make the similar samples more compact and the dissimilar samples more distant through the carefully designed contrastive loss. Meanwhile, the structural information of the label matrix is retained to make the model more robust and accurate.}
	\label{fig_framework}
\end{figure*}

The core idea of contrastive learning is to minimize the distance between feature representations of similar samples while maximizing the distance between feature representations of dissimilar samples. However, applying contrastive learning directly to multi-label classification is quite difficult. Because, unlike single-label classification, we can not identify positive and negative samples simply for an anchor point via its multiple labels. To bridge this gap, we design a contrastive loss for multi-label classification  from a) pulling together similar samples while enforcing low rank constraint, and b) pushing apart dissimilar samples while enforcing high rank constraint respectively. Figure \ref{fig_framework} illustrates the whole structure of the proposed method.

\subsubsection{\textbf{Pull together similar samples while enforcing low rank constraint}} 
Although we can not define absolute positive and negative samples according to the whole label set of a sample like single-label classification, we can find corresponding positive samples in each category in multi-label scenario. Specifically, for the i-th image  $\mathbf{x}_i$ with positive category $k$ in its label set, we define its positive sample set corresponding to the k-th category as $\mathbf{X}_{k}=\{ \mathbf{x}_{i}| y_{ik}^o=+1,\mathbf{x}_{i} \in \mathbf{X}\}$. At the same time, it can be found that we define the positive sample set based on a single \emph{category} rather than a single \emph{sample}, which is quite different from the traditional contrastive learning that requires defining positive and negative samples for an anchor point. 

Secondly, inspired by the classical multi-label classification methods \cite{7023448,8233207,7959114}, we choose to make similar samples more compact by rank constraint rather than distance. Therefore, for each category, we encourage the samples in the positive sample set of the category to be as compact as possible. Mathematically, we hope that in each category, the rank of the sub-matrix composed of deep embedding of samples sharing the same label should be as low as possible. The similarity encouraging item of CLML is designed as follows:
\begin{equation}
	 \sum ^{C}_{k=1}\left| \right| \mathbb{E} (\mathbf{X}_{k};\theta) \| _{\ast }
\end{equation}
Where $ \| \cdot \| _{\ast }$ denotes the sum of the singular values of a matrix, which is commonly used to strengthen the low-rank structure of a matrix.

\subsubsection{\textbf{Push apart dissimilar samples while enforcing high rank constraint}}
We get the positive sample set based on categories, but we can not get the negative sample set the same way. For example, let $\mathbf{x}_a,\mathbf{x}_b \in \mathbf{X}$, $y_{am}^o=+1,y_{bm}^o=-1$, and $y_{ap}^o=+1,y_{bp}^o=+1$. Even though $\mathbf{x}_a,\mathbf{x}_b$ are not in each other's positive sample set in the m-th category, they are in each other's positive sample set in the p-th category. Therefore, we further broaden our vision, no longer get the negative sample set from the just local perspective of category, but from the global perspective, take the entire samples as the negative sample set, and increase the distance between heterogeneous samples by imposing a high-rank constraint on the deep embedding matrix composed of the entire samples. In multi-label scenario, the labels of samples are diverse and often contain different labels, so each of these samples could more likely be negative samples of each other. By imposing a high-rank constraint on the deep embedding matrix composed of the entire samples, truly unrelated samples will be further apart, while strongly related samples will not be affected, thus more accurately pushing apart the heterogeneous samples. The following is the minimum expected loss term.
\begin{equation}
-\left\| \mathbb{E} \left( \mathbf{X};\theta \right) \right\| _{\ast }
\end{equation}

\subsubsection{\textbf{Contrastive loss}}
We increase the intra-class similarity and the inter-class separation by contrastive embedding. The contrastive loss function we designed is shown below:
\begin{equation}
	\mathcal{L}_{CL}:= \sum ^{C}_{k=1}\left| \right| \mathbb{E} (\mathbf{X}_{k};\theta) \| _{\ast }-\left\| \mathbb{E} \left( \mathbf{X};\theta \right) \right\| _{\ast }
\end{equation}
\begin{equation}
	\mathcal{L}_{CL}=\sum ^{C}_{k=1}\left| \right| \mathbf{Z}_{k} \| _{\ast }-\left\| \mathbf{Z} \right\| _{\ast }
	\label{optimize}
\end{equation}
Where $\mathbf{Z}_{k}=\mathbb{E}\left( \mathbf{X}_{k};\theta \right)$ is the sub-matrix of deep embedding belonging to label $k$.

Meanwhile, let $\mathbf{A}$, $\mathbf{B}$ and $\mathbf{C}$ be three matrices of $n$ row dimensions, and $[\mathbf{A}, \mathbf{C}]$ stands for the concatenation of $\mathbf{A}$ and $\mathbf{C}$, the same expression is used for $[\mathbf{A}, \mathbf{B}, \mathbf{C}]$ and $[\mathbf{B},\mathbf{C}]$. According to the relevant work of Ma et al. \cite{MA2021107675}, we can get the following formula:
\begin{equation}
\left\| \left[ \mathbf{A},\mathbf{B},\mathbf{C}\right] \right\| _{\ast }\leq \left\| \left[ \mathbf{A},\mathbf{C}\right] \right\| _{\ast }+\left\| \left[ \mathbf{B},\mathbf{C}\right] \right\| _{\ast }
\end{equation}

Therefore, it can be known that $ \sum ^{C}_{k=1}\left| \right| \mathbf{Z}_{k} \| _{\ast }$ is an upper bound of $\left\| \mathbf{Z} \right\| _{\ast }$, i.e., the contrastive loss we designed is always greater than or equal to 0. This means that our contrastive loss maintains the low-rank global structure. Therefore, we can find that our contrastive loss maintains a low-rank structure both globally and locally, which is exactly the characteristic of the label matrix \cite{MA2021107675}. In other words, our loss function makes the deep embedding matrix of samples inherit the structural characteristics of the label matrix, making the model more robust and accurate.

\subsection{Handling Missing Labels} 
\subsubsection{\textbf{The natural properties of CL}}
\label{nature}
In multi-label scenario with missing labels, finding missing labels using simply the feature representation that the model learns from the image is tricky. However, we notice that when images are similar, their label sets tend to be similar, and vice versa. This means that we can obtain more accurate feature representation by encouraging the reduction of image features' distance in samples with comparable label composition. That is the nature of our method. Our method imposes low-rank local label constraint on the matrix corresponding to each category, so that samples with similar labels are closer to each other even if the label set is not identical, indirectly narrowing the distance between false negative samples and true positive samples, thus better adapting to the scenario with missing labels.

Coincidentally, our method encourages the deep embedding matrix composed of all samples to be globally as high-rank as possible, providing less possibilities for the occurrence of missing labels, thus better detecting false negative labels.
\begin{figure}[ht]
	\centering
	\includegraphics[width=0.8\linewidth]{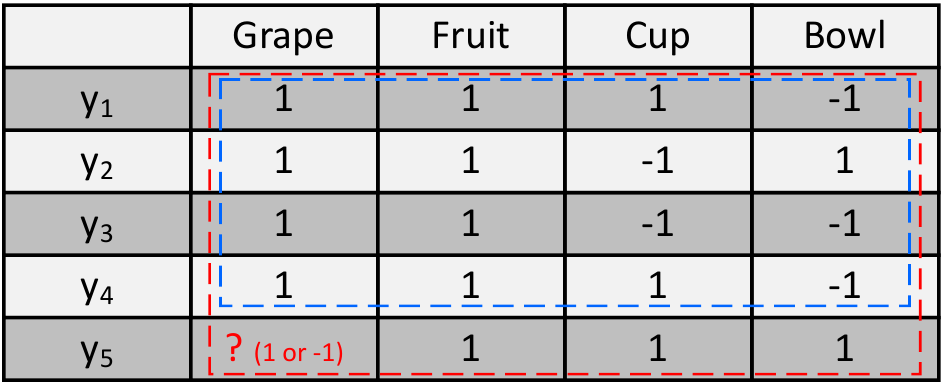}
	\caption{Samples of the shared ``Grape" are circled in blue, and the entire label matrix is circled in red. The rank of the matrix of the samples that share the ``Grapes" is less than the rank of the entire label matrix. Also, if we set the rank of the entire label matrix to 3 or 4, the missing values will be identified as 1 or -1. Obviously, a relatively high rank provides more possibilities for detecting missing labels.}
	\label{1or-1}
\end{figure}

Label structure preserving constrastive loss can relieve the problem of missing labels to some extent based on the two points mentioned above. To deal with the problem of missing labels even more effectively, we employ the concept of label correction to detect and update missing labels. In the following section, we will go over the specifics.

\subsubsection{\textbf{Update positive label set}}
\label{imbalance}
To further address the missing label problem, we introduce a label correction mechanism into our contrastive loss. The idea of this mechanism is partly inspired by the relevant work of Zhang et al. \cite{https://doi.org/10.48550/arxiv.2112.07368}.  Specifically, because the prediction of positive labels by neural network is very conservative, and the network is mnemonic, therefore it would be beneficial to utilize distinguishable missing labels early in model training rather than later. In the multi-label setting, the amount of negative labels in an image is generally considerably larger than the amount of positive labels, resulting in a serious positive and negative imbalance. This imbalance is particularly pronounced in the event of missing labels, interfering with the model's learning of positive samples. As a result of the small amount of positive labels, the model is cautious in predicting positive labels, and once the model determines that the label is positive, we can basically believe that it is correct. According to the memorization effect, ``deep neural networks (DNNs) learn simple patterns first and then memorize them." Zhang et al. \cite{https://doi.org/10.48550/arxiv.2112.07368} discovered that in the early stage of training, the model could distinguish missing labels from true negative labels based on the network's prediction probability, whereas in the later stage of training, the network tended to be over-fitting, making it difficult to distinguish the two. The experimental results are consistent with the theoretical conclusion of memorization effect. In light of the foregoing observations, we can conclude that missing labels can be effectively discovered and utilized in the early stage of model training according to its predicted probability, i.e., positive labels of samples can be regularly updated through the predicted probability of the model in the training stage. Thus, the positive label set with \textbf{La}bel \textbf{Co}rrection (LaCo) can be represented as follows:
\begin{equation}
\label{corL}
LaCo(y_{ij}^o)=\begin{cases}
	+1\ ,\ f\left( \mathbf{Z}_{i} \right)_{j} \geq \delta \\
	y_{ij}^o\ ,\ f\left( \mathbf{Z}_{i}\right)_{j}  <\delta \end{cases}
\end{equation}
Where $\delta$ is a proper threshold, and $\mathbf{Z}_{i}$ is the deep embedding of i-th image.

Therefore, by adding the label correction mechanism to Eq. (\ref{optimize}), CLML is rewritten as follows:
\begin{equation}
	\mathcal{L}_{CLML}=\sum ^{C}_{k=1}\left| \right| \left \{\mathbf{Z}_{i} | \widetilde{y}_{ik}=+1,\mathbf{Z}_i \in \mathbf{Z} \right\} \| _{\ast }-\left\| \mathbf{Z} \right\| _{\ast }
	\label{cl-mlml}
\end{equation}
\begin{equation}
	\widetilde{y}_{ik}=
	\begin{cases}
		y_{ij}^o\ ,\ N_e < N_E \\
		LaCo(y_{ij}^o)\ ,\ N_e \geq N_E 
	\end{cases}
\end{equation}
Where $N_e$ indicates the current epoch number, and $N_E \geq 1$ indicates that the epoch number of the positive label set starts updating. 

\subsection{Optimization}
Based on the work \cite{8578944}, we next describe how to optimize Eq. (\ref{optimize}) and Eq. (\ref{cl-mlml}) via backpropagation.
 
First, we need to compute the subgradient of nuclear norm $ \| \cdot \| _{\ast }$. To start with, let $\mathbf{A}=\mathbf{U}\Sigma \mathbf{V}$ be the singular value decomposition of the $N \times C$ matrix $\mathbf{A}$. Then set a smaller threshold $ \delta $, and assign the number of singular values greater than $ \delta $ to $s$. Then divide $\mathbf{U}$ and $\mathbf{V}$ as shown below:
\begin{displaymath}
	\mathbf{U}=\left[ \mathbf{U}^{1},\mathbf{U}^{2}\right]
\end{displaymath}
\begin{displaymath}
	\mathbf{V}=\left[ \mathbf{V}^{1},\mathbf{V}^{2}\right] 
\end{displaymath}
Where $\mathbf{U}^{1},\mathbf{V}^{1}$ are the first $s$ columns of $\mathbf{U}$ and $\mathbf{V}$, respectively. Therefore, a subgradient of the nuclear norm $\partial \left\| \mathbf{A}\right\| _{\ast }$ can be described as:
\begin{displaymath}
	\partial \left\| \mathbf{A}\right\| _{\ast }=\mathbf{U}^{1}{\mathbf{V}^{1}}^{T}
\end{displaymath}

With the above notations, for $\mathbf{X}_{k}$ the instance sub-matrix of each class $ k\in \left\{ 1,\ldots ,C\right\} $ in one minibatch, let $\mathbf{U}_{k}^{1},\mathbf{V}_{k}^{1}$ be the left and right singular vectors of $s$ columns, and $\mathbf{U}^{1},\mathbf{V}^{1}$ be the left and right singular vectors of $s$ columns of $\mathbf{X}$. Then the proposed following descent direction for Eq. (\ref{cl-mlml}) can be described as:
\begin{equation}
gradient \ \mathcal{L}_{CL}(\mathbf{X}):=\sum_{k}^{C}\left[\mathbf{R}_{k}^{(l)}\left|\mathbf{U}_{k}^{1} \mathbf{V}_{k}^{1}\right| \mathbf{R}_{k}^{(r)}\right]-\mathbf{U}^{1} \mathbf{V}^{1^{T}}
\label{gradient}
\end{equation}
Here, $\mathbf{R}_{k}^{\left( l\right) },\mathbf{R}_{k}^{\left( r\right) } $ are all-zero matrices to complete the dimensions of $\mathbf{X}$. The first part in Eq. (\ref{gradient}) encourages the low-rank structure of per-label matrix in embedding space, thus reducing the variance of primary components of per-class instances. The second part globally encourages a maximally separated structure for the whole label matrix in embedding sapce, thus increasing the variance of all the instances.

\subsection{Model Learning}
To the best of our knowledge, CLML is the first general multi-label CL loss in the missing-label scenario. It can seamlessly be paired with those losses of any existing multi-label learning methods just via a single hyperparameter. Specifically, we seek to minimize the following objective function:
\begin{equation}
	\min _{\theta, \phi }\mathcal{L}_{classification}\left( \mathbf{X},\mathbf{Y}^o,\theta, \phi \right) +\lambda \cdot \mathcal{L}_{CLML} \left( \mathbf{X},\mathbf{Y}^o,\theta \right)
	\label{lam7}
\end{equation}
Here $\lambda$ is a hyper-parameter trading off the existing loss $\mathcal{L}_{classification}$ and contrastive loss. $ \theta$ refers to parameters in the embedding model. $\mathbf{Y}^o$ is the label matrix. In order to show the details of the proposed method more intuitively, we show the model learning process in Algorithm \ref{alg1}.

\begin{algorithm}
	\renewcommand{\algorithmicrequire}{\textbf{Input:}}
	\renewcommand{\algorithmicensure}{\textbf{Output:}}
	\caption{Label Structure Preserving Contrastive Embedding}
	\label{alg1}
	\begin{algorithmic}[1]
	\REQUIRE
	Training data matrix $\mathbf{X}$, label matrix $\mathbf{Y}^o$, deep embedding network $\mathbf{Z} = \mathbb{E}(\mathbf{X}, \theta)$ and deep multi-label classifier $f(\mathbf{Z}, \phi)$, trade-off parameter $\lambda$.
	\ENSURE The well trained deep model $\mathbb{E}(\cdot, \theta)$ and $f(\cdot,\phi)$
		\FOR{$N_e = \left \{1,\cdots,N_{epoch}\right \}$}
		\FOR{each minibatch $\mathbf{X}_b$ and $\mathbf{Y}^o_b$}
		\STATE $\mathbf{Z}_b = \mathbb{E}(\mathbf{X}_b, \theta)$
		\STATE Correct the false negative labels in $\mathbf{Y}^o_b$ according to Eq. (\ref{corL})
		\STATE Calculate contrastive loss $\mathcal{L}_{CL}(\mathbf{Z}_b,\mathbf{Y}^o_b)$ according to Eq. (\ref{cl-mlml})
		\STATE Calculate total loss according to Eq. (\ref{lam7})
		\STATE Backpropagation
		\ENDFOR

		\ENDFOR
	\end{algorithmic}  
\end{algorithm}

\section{Experiment}
In this section, we demonstrate the improved performance when using a combination of CLML and existing loss in several different standard visual classification datasets by comparing various loss functions.


\subsection{Experimental Setup}
\subsubsection{\textbf{Datasets}}
Our model is tested on three multiple standard multi-label benchmarks: MS-COCO (COCO) \cite{10.1007/978-3-319-10602-1_48}, PASCAL VOC 2012 (VOC) \cite{bb32678ecd9b4bc296dd558806cda288}, and NUS-WIDE (NUS) \cite{10.1145/1646396.1646452}. 
\begin{enumerate}[(1)]
	\item MS-COCO: COCO is a large-scale dataset, in which the images include natural images and common target images in life, with complex backgrounds and a large number of targets. It contains 122,218 images with 80 classes, among which the training set consists of 82,081 images and the validation set of 40,137 images.
	\item PASCAL VOC 2012: VOC is one of the benchmarks frequently used in image classification comparison experiments and model evaluation. It contains 5717 training images with 20 classes and 5823 test images.
	\item NUS-WIDE: NUS is a large multi-label dataset containing 81 classes. We collected 119,103 and 50,720 images for our training set and test set.
\end{enumerate}

To construct a training setting of missing labels, in the COCO dataset, we randomly discard the labels in the training set based on the retention ratios of 75\% labels, 40\% labels, and single label, whereas in the VOC and NUS datasets, we discard the labels in the training set based on the retention ratio of single label. The following Table \ref{tab:freq} describes the specific settings of the datasets.
\begin{table}[!h]
\caption{Specific data from different datasets}
\centering
\label{tab:freq}
\begin{tabular}{lcccc}
		\toprule
		& samples & classes & labels  & avg.label/img \\
		\midrule
		COCO-full labels      & 82,081  & 80      & 241,035 & 2.9           \\
		COCO-75\% labels left & 82,081  & 80      & 181,422 & 2.2           \\
		COCO-40\% labels left & 82,081  & 80      & 96,251  & 1.2           \\
		COCO-single label     & 82,081  & 80      & 82,081  & 1.0           \\
		\midrule
		NUS-full labels       & 119,103 & 81      & 289,460 & 2.4           \\
		NUS-single label      & 119,103 & 81      & 119,103 & 1.0           \\
		\midrule
		VOC-full labels       & 5,717   & 20      & 8,331   & 1.4           \\
		VOC-single label      & 5,717   & 20      & 5,717   & 1.0           \\
		\bottomrule
\end{tabular}
\end{table}
\subsubsection{\textbf{Evaluation metrics}}
Mean average precision (mAP) is widely used to evaluate model performance, and a higher value $\uparrow$ indicates better model performance. We also included the overall F1-measure (OF1), per-category F1-measure (CF1) and their corresponding Precision (P) and Recall (R) into our evaluation indexes. In addition, we set the confidence threshold at 0.5, which means that if the predicted probability of a candidate label is more than 0.5 during model testing, it will be regarded as positive; otherwise, it will be regarded as negative
\subsubsection{\textbf{Baselines}}
\label{baselines}
CLML is the first general multi-label CL loss in the missing-label scenario. It can seamlessly be paired with those losses of any existing multi-label learning methods just via a single hyperparameter $\lambda$ in Eq.(\ref{lam7}). Through certain experimental attempts, we found that $\lambda=1$ has better performance in most cases. For the convenience of the experiments, we set $\lambda=1$ constantly. In the following experiments, we compare the classical BCE loss, Focal loss \cite{8237586} and the recently proposed state-of-the-art loss Hill \cite{https://doi.org/10.48550/arxiv.2112.07368} and SPLC \cite{https://doi.org/10.48550/arxiv.2112.07368} in MLML tasks. The comparison models' results are derived from the best-reports in their respective articles. 

\begin{table*}[h] 
\scriptsize
\centering
\caption{Compared results on COCO dataset with varied missing label ratios}
\label{com}
\begin{tabular}{cc|c|cc|cc|cc|cc}
		\toprule
		\multicolumn{2}{c|}{Method}              & \begin{tabular}[c]{@{}c@{}}BCE\\ (full labels)\end{tabular} & \multicolumn{1}{c}{BCE} & \multicolumn{1}{c|}{BCE+CLML} & \multicolumn{1}{c}{Focal \cite{8237586}} & \multicolumn{1}{c|}{Focal+CLML} & \multicolumn{1}{c}{Hill \cite{https://doi.org/10.48550/arxiv.2112.07368}} & \multicolumn{1}{c|}{Hill+CLML} & \multicolumn{1}{c}{SPLC \cite{https://doi.org/10.48550/arxiv.2112.07368}} & \multicolumn{1}{c}{SPLC+CLML} \\
		\midrule
		\multirow{7}{*}{75\% labels left} & mAP $\uparrow$ & 80.3                               & 76.8                    & \textbf{78.0}              & 77.0                      & \textbf{78.3}                & 78.8                     & \textbf{79.6}               & 78.4                     & \textbf{80.4}               \\
		& CP$\uparrow$  & 80.8                                & 85.1                    & \textbf{86.2}              & 83.8                      & \textbf{86.0}                & \textbf{73.6 }                    & 72.8             & 72.6                     & \textbf{75.6}               \\
		& CR$\uparrow$  & 70.3                                  & 58.1                    & \textbf{58.7}              & 59.4                      & \textbf{61.0}                & 74.4                     & \textbf{76.3}               & \textbf{75.1}            & 74.6                        \\
		& CF1$\uparrow$ & 74.9                                & 67.7                    & \textbf{68.5}              & 68.4                      & \textbf{69.7}                & 73.6                     & \textbf{74.1}               & 73.2                     & \textbf{74.8}               \\
		& OP$\uparrow$  & 84.3                                & 90.1                    & \textbf{90.9}              & 88.6                      & \textbf{89.1}                & \textbf{76.4}            & 74.6                       & 74.0                     & \textbf{79.1}               \\
		& OR$\uparrow$  & 74.2                               & 58.7                    & \textbf{59.3}              & 59.8                      & \textbf{61.2}                & 78.3                     & \textbf{80.3}               & \textbf{79.3}            & 78.0                        \\
		& OF1$\uparrow$ & 78.9                                & 71.1                    & \textbf{71.8}              & 71.4                      & \textbf{72.6}                & 77.3                     & 77.3             & 76.6                     & \textbf{78.5}               \\
		\midrule
		\multirow{7}{*}{40\% labels left} & mAP$\uparrow$ & -                                                           & 70.5                    & \textbf{71.5}              & 71.7                      & \textbf{73.3}               & 75.2                     & \textbf{76.4}               & 75.7                     & \textbf{76.5}               \\
		& CP$\uparrow$  & -  & 89.2          & \textbf{89.5}                      & 88.9&\textbf{ 89.2} & 80.4 & \textbf{81.2}               & \textbf{81.6}            & 79.6                        \\
		& CR$\uparrow$  & -                                                           & 34.4                    & \textbf{36.0}              & 37.0                      & \textbf{40.3}                & 61.4                     & \textbf{62.6}               & 60.7                     & \textbf{65.8}               \\
		& CF1$\uparrow$ & -                                                           & 45.8                    & \textbf{47.8}              & 48.7                      & \textbf{52.1}                & 68.6                     & \textbf{69.6}               & 67.9                     & \textbf{70.8}               \\
		& OP$\uparrow$  & -                                                           & \textbf{94.1}           & 93.7                       & \textbf{93.7}             & 93.3                         & 85.5                     & \textbf{96.5}               & \textbf{87.7}            & 83.6                        \\
		& OR$\uparrow$  & -                                                           & 25.7                    & \textbf{27.3}              & 28.8                      & \textbf{31.2}                & 63.8                     & \textbf{64.6}               & 63.0                     & \textbf{69.5}               \\
		& OF1$\uparrow$ & -                                                           & 40.4                    & \textbf{42.3}              & 44.0                      & \textbf{47.0}                & 73.1                     & \textbf{74.0}               & 73.3                     & \textbf{75.9}               \\
		\midrule
		\multirow{7}{*}{single label}     & mAP$\uparrow$ & -                                                           & 68.6                    & \textbf{69.5}              & 70.2                      & \textbf{71.8}                & 73.2                     & \textbf{74.0}               & 73.2                     & \textbf{74.0}               \\
		& CP$\uparrow$  & -                                                           & 88.6                    & \textbf{89.1}              & 88.2                      & \textbf{88.9}                & 79.7                     & \textbf{83.0}               & \textbf{83.8}            & 80.9                        \\
		& CR$\uparrow$  & -                                                           & 33.0                    & \textbf{33.5}              & 36.0                      & \textbf{37.4}                & \textbf{58.0}              & 55.7                       & 53.1                     & \textbf{58.7}               \\
		& CF1$\uparrow$ & -                                                           & 43.8                    & \textbf{44.2}              & 47.0                      & \textbf{48.6}                & \textbf{65.5}            & 64.2                       & 61.6                     & \textbf{65.5}               \\
		& OP$\uparrow$  & -                                                           & 93.9                    & \textbf{94.8}              & 93.4                      & \textbf{93.9}                & 85.3                     & \textbf{88.7}               & \textbf{90.1}            & 86.4                        \\
		& OR$\uparrow$  & -                                                           & 23.6                    & \textbf{24.5}              & 26.6                      & \textbf{28.3}                & \textbf{58.7}            & 55.0                      & 53.8                     & \textbf{60.5}               \\
		& OF1$\uparrow$ & -                                                           & 37.7                    & \textbf{38.9}              & 41.4                      & \textbf{43.5}                & \textbf{69.5}            & 67.8                     & 67.4                     & \textbf{71.2}               \\
		\bottomrule
\end{tabular}
\end{table*}
\subsubsection{\textbf{Implementation details}}
For feature extraction, we utilize the Resnet101 pretained on ImageNet \cite{inproceedings} as the backbone network. The input images are resized to $448 \times 448$. We choose Adam as the optimizer for our model. The initial learning rate is 1e-4. To improve the model's robustness, the EMA method is utilized to average the model's parameters. In addition, we leverage data augment to improve the model's robustness, which is inspired by contrastive learning. Furthermore, all experiments are run on 2 Nvidia RTX A6000 GPU with 48GB using a deep learning toolbox PyTorch. Then, our model is trained for 30 epochs on MSCOCO and NUS with a batch-size of 128, and 80 epochs on VOC with a batch-size of 128.
\subsection{Experimental Results on MSCOCO}
\subsubsection{\textbf{Classification results}}
\begin{figure}[ht]
	\centering
	\includegraphics[width=0.8\linewidth]{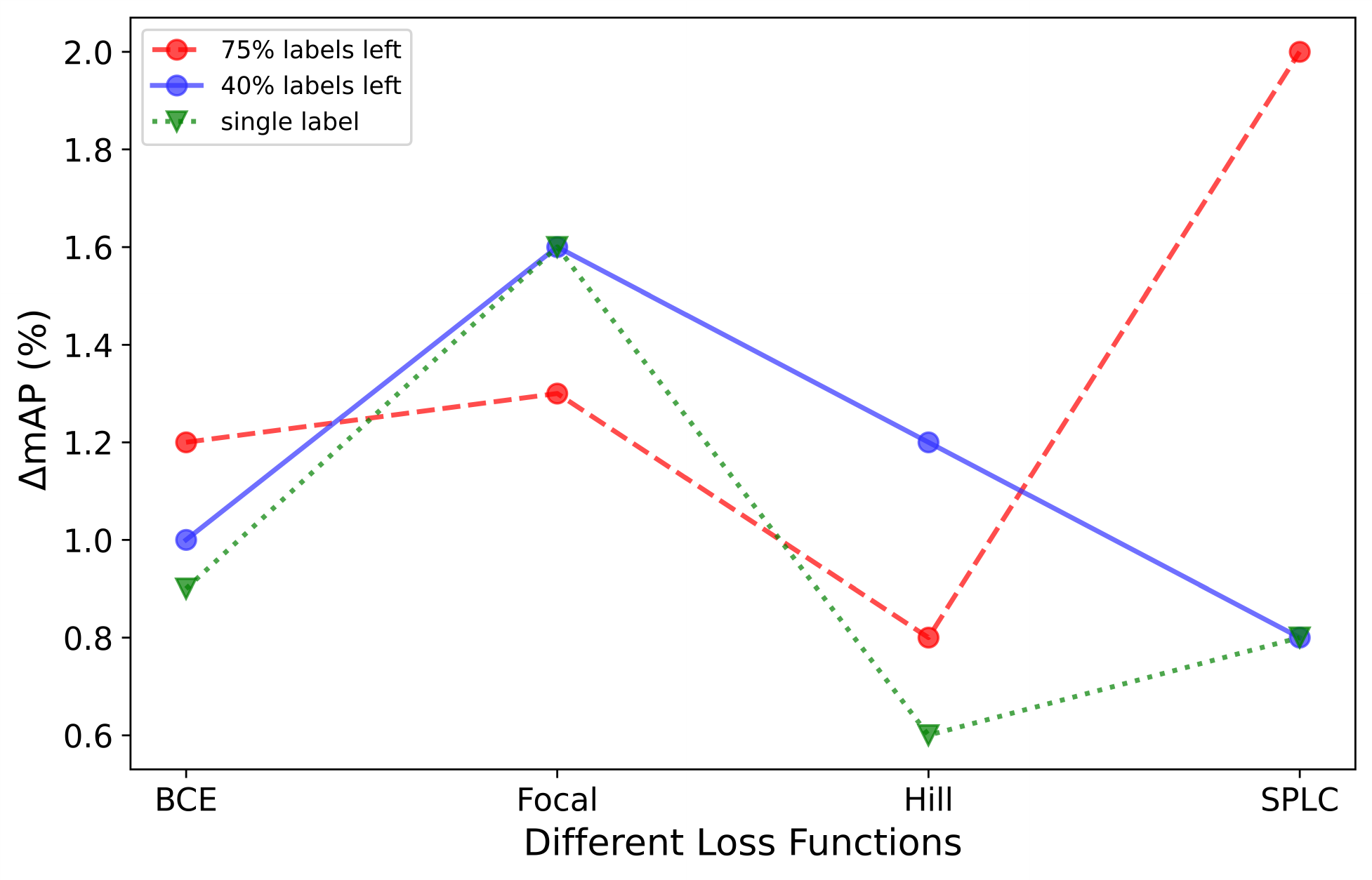}
	\caption{The increase amplitude of CLML to different loss functions on training set with different ratios of missing labels. }
	\label{fig:delta}
\end{figure}

We first conduct experiments on the COCO dataset. It is worth noting that COCO dataset is a challenging dataset for image classification due to the complex background of images. Therefore, in order to verify that the CLML can effectively improve the performance of the model, we use a series of evaluation metrics to testify the performance of the model. The comparsion results are shown in Table \ref{com}. Meanwhile, Figure \ref{fig:delta} shows how the amplitude of different loss functions increases with CLML on a training set with varied missing label ratios. According to the experimental results, we have the following observations.

\begin{figure}[ht]
	\centering
	\includegraphics[width=0.8\linewidth]{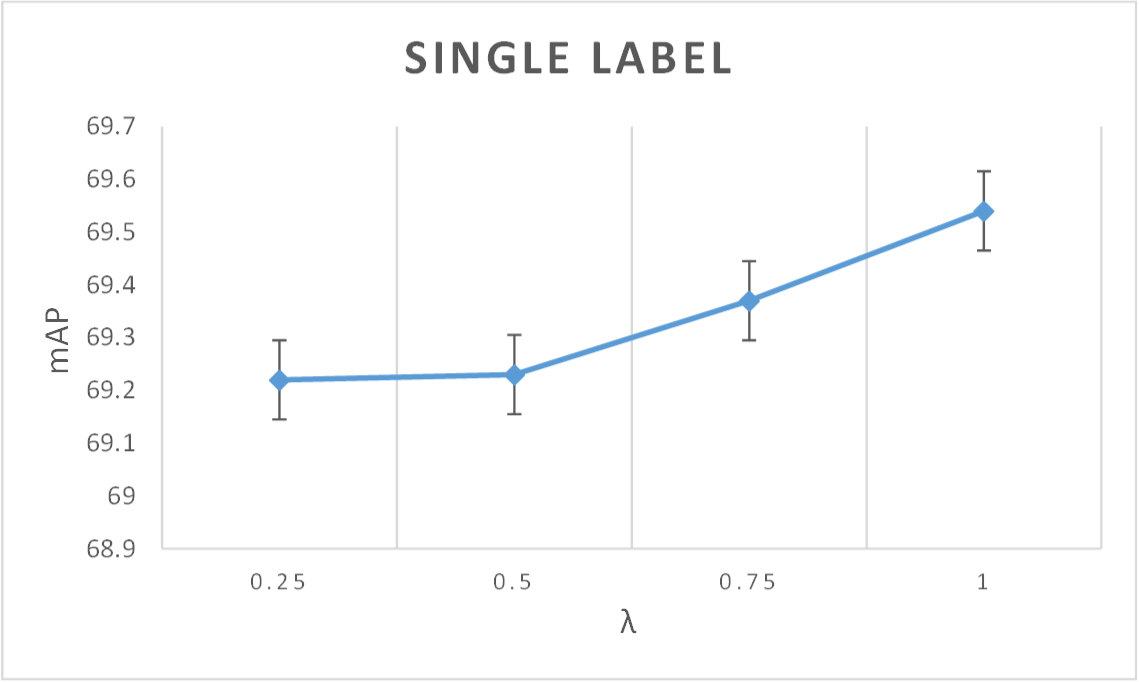}
	\caption{In the case of single label, the influence of different $\lambda$ on model performance.}
    \label{fig:lambda}
\end{figure}

\begin{enumerate}[(1)]
	\item In three training settings with varying degrees of missing label, the classification performance of loss function with CLML is better than its own on almost all evaluation metrics, which proves the effectiveness of CLML.

	\item As shown in Figure \ref{fig:delta}, although CLML improves all loss functions in COCO dataset to some extent, the improvement of loss functions in training set with 75\% and 40\% left labels is generally better than that of single label. This is because our model can capture less label information for the dataset with only one label, therefore the promotion of loss functions with CLML is not as good as it is for the other two training sets. However, CLML can increase the model's classification performance by increasing similarity within classes and separation between classes, which is why $ \Delta mAP$ in Figure \ref{fig:delta} is always greater than 0.
	
	\item Notably, it can be found that SPLC loss combined with CLML can achieve mAP of 80.4\%, and it is even better than that of BCE loss training on fully labeled dataset, which could not be achieved by SPLC alone. This is because, compared with BCE loss, SPLC loss paired with CLML not just alleviates the category imbalance problem discussed in Section \ref{imbalance}, but also improves the model's discriminant power through contrastive learning. 

	\item As can be shown in Figure \ref{fig:lambda}, in the extreme case of single label only, we can see that when $\lambda =1$, the model has the best performance. Therefore, for the sake of simplicity, $\lambda =1$ is used uniformly in the following experiments.
	
\end{enumerate}

\subsubsection{\textbf{Take advantage of the relevance of labels}}
As mentioned in Section \ref{nature}, our model can detect false negative labels by similar label composition, i.e., taking advantage of strong correlation between labels. Therefore, in order to verify that our model effectively utilizes the correlation of labels, we use Apriori algorithm \cite{10.5555/645920.672836} to mine the association rules of COCO training set, where we set the minimum support to 0.001 and the minimum confidence to 0.8. The category groups with strong label correlation in the COCO training set are shown in the Table \ref{rules}.

\begin{table}[!h] 
\scriptsize
\centering
\caption{Some association rules in the COCO training set. Conf refers to the confidence of the association rules.}
\label{rules}
\begin{tabular}{|c|c|c|}
		\hline
		category group                                                                                                                  & association rules                                        & conf \\ \hline
		\multirow{3}{*}{\begin{tabular}[c]{@{}c@{}}bowl, bottle, cake, \\ chair, cup, pizza, \\ sandwich, dining table\end{tabular}} & chair, cup, pizza, bottle $\rightarrow$ dining table & 0.96       \\ \cline{2-3} 
		& chair, cup, bowl, cake $\rightarrow$  dining table   & 0.97       \\ \cline{2-3} 
		& sandwich, cup, bowl $\rightarrow$  dining table      & 0.91       \\ \hline
		\begin{tabular}[c]{@{}c@{}}book, keyboard, \\ laptop, mouse\end{tabular}                                                     & keyboard, laptop, book $\rightarrow$ mouse           & 0.86       \\ \hline
\end{tabular}
\end{table}

\begin{figure*}[htbp]
\centering
\includegraphics[width=\textwidth]{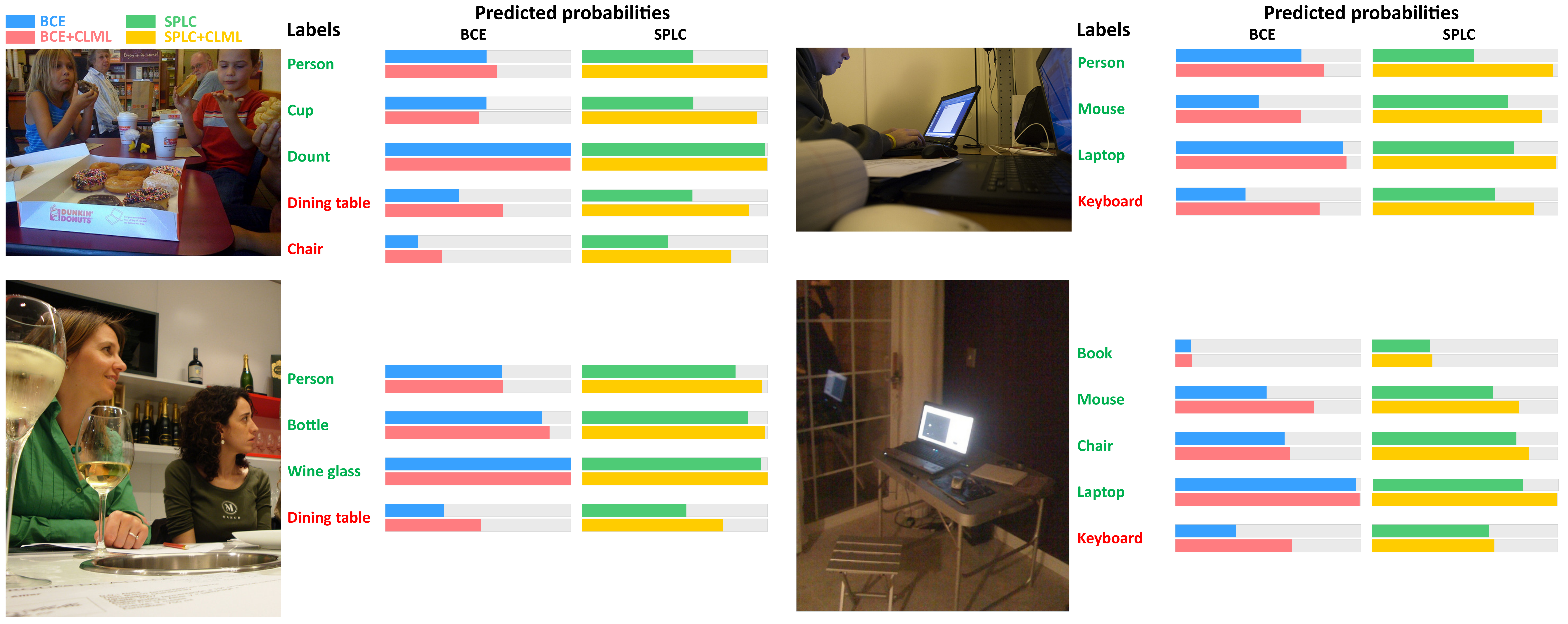}
\caption{On the COCO validation set, we compare BCE loss and SPLC loss with and without CLML. Positive labels and missing labels are represented by green and red, respectively. For the predicted probabilities, blue and pink represent BCE loss with and without CLML, respectively. Green and orange represent SPLC loss with and without CLML, respectively. The results show that loss with CLML can better predict missing labels via label correlation, such as the strong association between food and dining table, as well as the strong correlation between laptop, mouse, and keyboard.}
\label{result_pred}
\end{figure*}

\begin{figure}[!ht] 
\centering
\includegraphics[width=.9\linewidth]{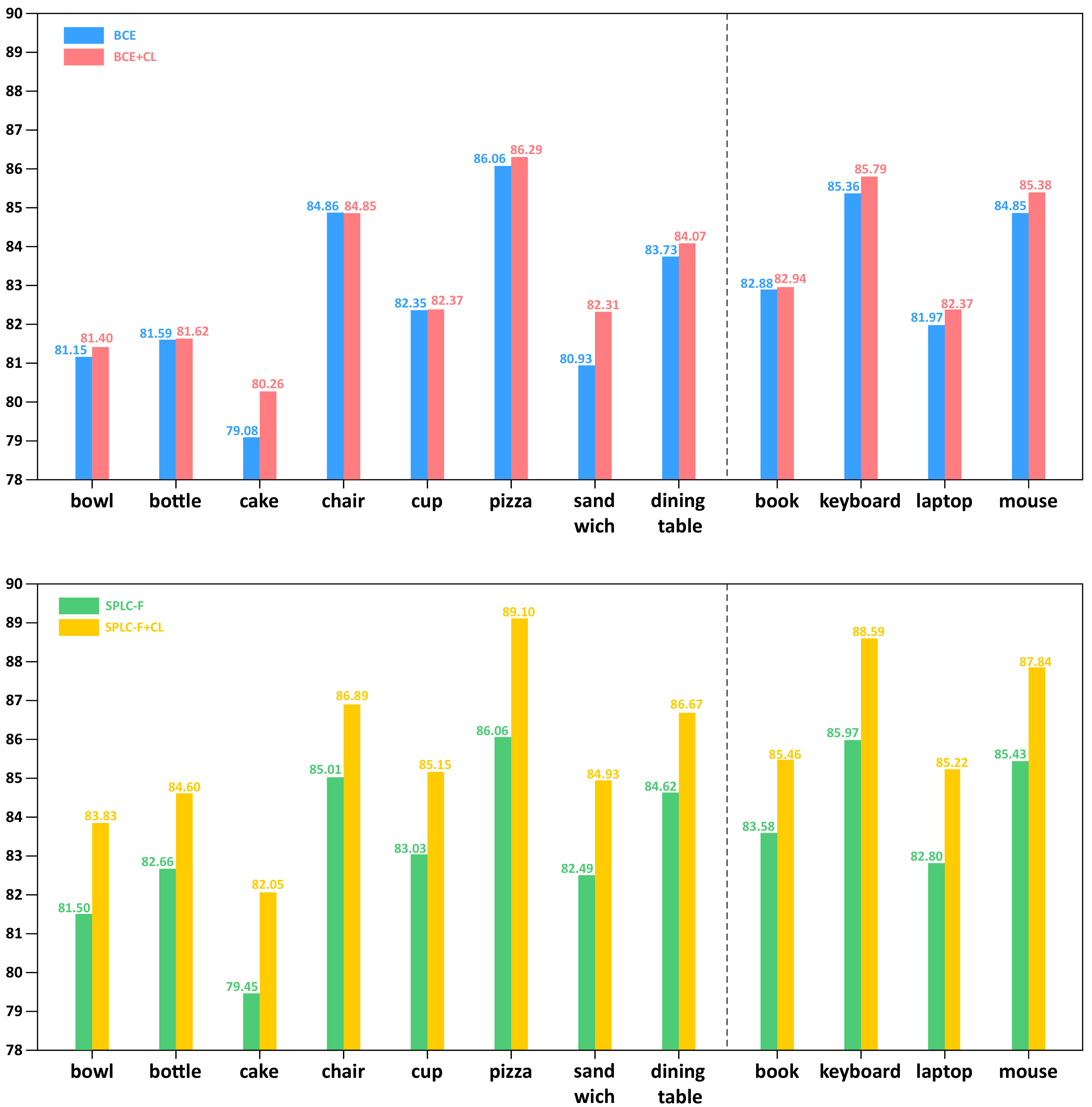}
\caption{AP scores of four loss functions in different categories. The classification performance of the BCE loss with and without CLML is on the upper side, and the classification performance of the SPLC loss with and without CLML is on the lower side.}
\label{fig:rule}
\end{figure} 
From the results of the Table \ref{rules}, we can see that when the items labeled as ``bowl", ``bottle", ``cake", ``chair", ``cup", ``pizza" and ``sandwich" appear in a certain combination, the items labeled as ``dining table" also appear in a high frequency. Therefore, we believe that there is a strong correlation between these labels. The same goes for ``book", ``keyboard", ``laptop", and ``mouse". Then, we compare the average precision (AP) scores of the above labels on COCO validation set of BCE loss and SPLC loss with and without CLML. Blue, pink, green and orange represent the four loss functions respectively, and the results are shown in Figure \ref{fig:rule}.

In order to verify the improvement brought by CLML to the model, we select the basic BCE loss and SPLC loss, which achieves state-of-the-art loss functions in MLML, as the comparison loss functions. Figure \ref{fig:rule} depicts the AP scores for the two above-mentioned category groups using the two loss functions, with dotted lines separating the two groups of labels. By observing subfigure in the upper side of Figure \ref{fig:rule}, we can find that for the first category group, compared with only BCE, AP scores of most categories are improved to varying degrees with CLML. In the second category group, BCE loss with CLML yields higher AP scores on all the 4 categories, with the AP scores of mouse improving by 0.53\% at most, while that of book improving by only 0.06\%. Because the scene in which the book appears is relatively simple in the COCO dataset, primarily involving a laptop, keyboard, and mouse, it is more likely for the mouse to appear in such a situation, and the reverse is not necessarily true. This phenomena also conforms to the previously mentioned label asymmetry co-occurrence relation. The same conclusion can be drawn from the the lower subgraph of Figure \ref{fig:rule}.

\subsubsection{\textbf{Predicted probabilities}}
Also, to verify the efficacy of CLML in processing images with missing labels, we give examples of predicted probabilities from BCE, and SPLC with and without CLML in Figure \ref{result_pred}. Among them, the green labels are the ground truth of the images, and the red labels are the labels missed by human annotators but recognized by our model. It can be found that the BCE loss with CLML is far superior to the BCE loss in the ability of mining missing labels in images, and this ability is particularly evident in images with strong correlation labels, such as food and dining table, laptop, mouse and keyboard. Moreover, on the basis of the SPLC, the improvement is even more striking.

\begin{figure*}[htbp] 
\centering
\includegraphics[width=\linewidth]{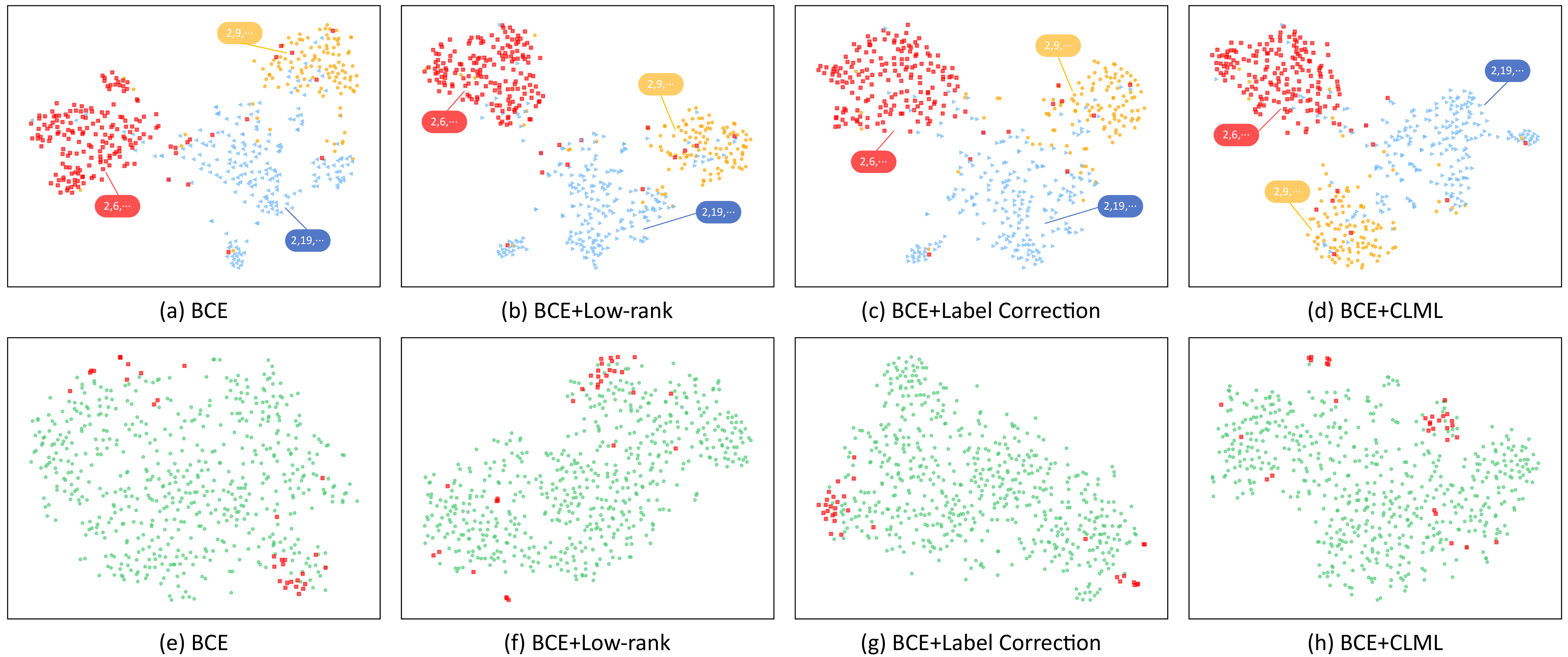}
\caption{t-SNE visualization for the learned embedding features of the multi-label images on VOC 2012 test set. Subfigures (a) to (h) are generated by ResNet101 and its corresponding loss function respectively. For subfigures (a) to (d), different categories are indicated by different colors, shapes, and numbers, respectively. For subfigures (e) to (h), green represents the true positive labels, and red represents the missing labels, i.e., the false negative labels.}
\label{fig_tsne}
\end{figure*}
\subsection{Experimental Results on Pascal VOC 2012 and NUS-WIDE}
\subsubsection{\textbf{Classification results}}
\label{class}
We also conduct our experiment on VOC and NUS datasets. Just like COCO dataset, we report results on a series of evaluation metrics, as shown in Table \ref{voc-nus}. 
\begin{table}[!h]
\centering
\caption{Compared result on VOC and NUS datasets}
\label{voc-nus}
\begin{tabular}{l|ccc|ccc}
		\toprule
		\multicolumn{1}{c|}{\multirow{2}{*}{Method}} & \multicolumn{3}{c|}{VOC-single label} & \multicolumn{3}{c}{NUS-single label} \\
		\multicolumn{1}{c|}{}       & mAP$\uparrow$    & CF1$\uparrow$    & OF1$\uparrow$   & mAP$\uparrow$    & CF1$\uparrow$    & OF1$\uparrow$   \\
		\midrule
		BCE\footnotesize(full labels)   & 89.0     & 83.3    & 85.6  & 60.6 & 59.1     & 73.6 \\
		BCE                          & 85.6   & 77.4   & 78.4  & 51.7   & 30.9 & 33.3 \\
		BCE+CLML                      &\textbf{87.2}   & \textbf{79.3}& \textbf{80.3}&\textbf{53.0} &\textbf{ 31.1} &\textbf{34.6}\\
		\midrule
		Focal \cite{8237586}     &86.8& 78.2 &79.1&53.6&34.2&35.0    \\
		Focal+CLML                  &\textbf{87.5}&\textbf{79.8}&\textbf{80.3} &\textbf{54.7}  &  \textbf{35.0} &\textbf{36.5}  \\
		\midrule
		Hill  \cite{https://doi.org/10.48550/arxiv.2112.07368} & 87.8&81.1& \textbf{83.8 } & 55.0  &\textbf{54.1}  &\textbf{68.6} \\
		Hill+CLML        &\textbf{88.1}&\textbf{81.5}& 82.0      &\textbf{55.4} & 51.2   & 65.6  \\
		\midrule
		SPLC \cite{https://doi.org/10.48550/arxiv.2112.07368} & \textbf{88.1} &80.2&\textbf{83.0}  & 55.2   &\textbf{52.4} &\textbf{70.6  }\\
		SPLC+CLML            & 88.0 &\textbf{81.5} &82.4 &\textbf{55.3} & 48.6 & 61.0\\
		\bottomrule
\end{tabular}
\end{table}

From Table \ref{voc-nus}, we have the following observations.
\begin{enumerate}[(1)]
	\item In VOC and NUS datasets with single label, the loss functions with CLML have been improved in most evaluation metrics, which proves the effectiveness of the proposed method.
	\item CLML has obvious advantages in the classic BCE and Focal, whereas in the Hill and SPLC, CLML seems willing but weak. This could be because the class imbalance is more prominent in the VOC and NUS datasets with only one label. Therefore, Hill and SPLC, which are generated in a design for class imbalance, leave CLML with very little space to play.
\end{enumerate}

\subsection{Ablation Study}
\label{bce-sne}
In the method of multi-label classification using the BCE loss function, the fully connected layer of the last layer of the model establishes a corresponding linear classifier for each label. However, in this way, each linear classifier pays more attention to specific labels and less attention to the discrimination of features belonging to different labels. However, in multi-label learning, the strength of model discrimination ability largely determines its classification accuracy. Our model uses contrastive learning to compensate for the BCE loss function's lack of model discrimination ability. To prove this and explore how CLML improves the performance of the model, we conduct an ablation experiment based on BCE loss.

First of all, we conduct ablation experiments on VOC datasets with single label. BCE and BCE combined with low-rank constraint, namely Eq. (\ref{optimize}), BCE combined with label correction, and BCE combined with CLML, are used to train the model. The specific classification results are shown in Table \ref{table::bce-map}. From the classification results, we can see that CLML has obtained the optimal or sub-optimal value in all evaluation indexes. 

Simultaneously, to more vividly explain the role of each component of CLML in improving model performance, we discuss and analyze the improvement of intra-class similarity and inter-class separation, and the narrowing of the distance between false negative samples and true positive samples. For the former, we select three groups of similar scenes containing the same label for t-SNE feature visualization, as shown in subfigure (a) and subfigure (d) in Figure \ref{fig_tsne}. For the latter, we select one category in VOC test set to observe the distribution of positive and false negative samples in the embedding space under different loss function backgrounds,as shown in subfigure (e) and subfigure (h) in Figure \ref{fig_tsne}. 

\begin{table}[!ht] 
\scriptsize
\centering
\caption{Ablation study of different loss functions. The best value is in bold and the second best value is underlined}
\label{table::bce-map}
\begin{tabular}{|c|ccccccc|}
	\hline
	\multirow{2}{*}{Method}                                             & \multicolumn{7}{c|}{VOC-single label}                 \\ \cline{2-8} 
	& \multicolumn{1}{c|}{mAP$\uparrow$}           & \multicolumn{1}{c|}{CP$\uparrow$}            & \multicolumn{1}{c|}{CR$\uparrow$}            & \multicolumn{1}{c|}{CF1$\uparrow$}           & \multicolumn{1}{c|}{OP$\uparrow$}            & \multicolumn{1}{c|}{OR$\uparrow$}            & OF1$\uparrow$           \\ \hline
	\begin{tabular}[c]{@{}c@{}}BCE\\ only\end{tabular}                  & \multicolumn{1}{c|}{85.6}          & \multicolumn{1}{c|}{87.8}          & \multicolumn{1}{c|}{71.6}          & \multicolumn{1}{c|}{77.4}          & \multicolumn{1}{c|}{91.1}          & \multicolumn{1}{c|}{68.7}          & 78.4          \\ \hline
	\begin{tabular}[c]{@{}c@{}}BCE with\\ Low-rank\end{tabular}         & \multicolumn{1}{c|}{{\ul 87.0}}    & \multicolumn{1}{c|}{\textbf{91.2}} & \multicolumn{1}{c|}{70.8}          & \multicolumn{1}{c|}{78.3}          & \multicolumn{1}{c|}{\textbf{93.5}} & \multicolumn{1}{c|}{69.7}          & 79.9          \\ \hline
	\begin{tabular}[c]{@{}c@{}}BCE with\\ Label Correction\end{tabular} & \multicolumn{1}{c|}{86.8}          & \multicolumn{1}{c|}{89.5}          & \multicolumn{1}{c|}{{\ul 72.2}}    & \multicolumn{1}{c|}{{\ul 78.5}}    & \multicolumn{1}{c|}{92.1}          & \multicolumn{1}{c|}{\textbf{72.1}} & \textbf{80.9} \\ \hline
	\begin{tabular}[c]{@{}c@{}}BCE with\\ CLML\end{tabular}             & \multicolumn{1}{c|}{\textbf{87.2}} & \multicolumn{1}{c|}{{\ul 90.4}}    & \multicolumn{1}{c|}{\textbf{72.4}} & \multicolumn{1}{c|}{\textbf{79.3}} & \multicolumn{1}{c|}{{\ul 92.2}}    & \multicolumn{1}{c|}{{\ul 71.1}}    & {\ul 80.3}    \\ \hline
\end{tabular}
\end{table}

\subsubsection{\textbf{Effectiveness of low-rank constraint}} Let's focus on the subgraphs (a), (b), (e) and (f) in Figure \ref{fig_tsne}. We can find that the clusters in subfigure (b) are more compact and the boundaries of the three clusters are clearer, which show that low-rank constraint effectively improves the intra-class similarity and inter-class separation. By comparing subfigures (e) and (f), we can also find that the sample distribution in subfigure (f) is more compact, which is consistent with the conclusion obtained by comparing subfigures (a) and (b). Meanwhile, from subfigure (e), in the case of only BCE loss guidance model training, false negative samples are mainly distributed in the edges of the real category, and from the subfigure (f), we can see that false negative samples are closer to the center of the real category in the case of joining the low-rank constraint guidance model training, which corresponds to the fact that CLML naturally mitigates missing labels mentioned in Section \ref{nature}. 

\subsubsection{\textbf{Effectiveness of label correction}}

Then, let's focus on the four subgraphs (a), (c), (e) and (g) in Figure \ref{fig_tsne}. It can be found from subfigures (c) that features in the embedding space become looser after adding label correction. This is because in the process of model training, the missing labels of samples are constantly replenished, and samples will belong to more categories, thus making features looser. This phenomenon strongly indicates that our model mines missing labels.

\subsubsection{\textbf{The advantage of CLML}}

Finally, we compare the subgraphs (d) and (h) with the rest of the subgraphs in Figure \ref{fig_tsne}. Compared with the subgraph (b), the features of the samples in subgraph (d) are looser, because it mines more missing labels. Compared with subgraph (c), the features of the samples in subgraph (d) are more compact, because it narrows the distance between similar samples. Meanwhile, in subgraph (d), the boundary of each cluster is more clearer, because it expands the distance between dissimilar samples.
By comparing the subgraphs (f) and (h), similarly, the sample features in (h) are looser, suggesting that missing labels have been detected. By comparing subgraphs (g) and (h), we can obviously see that more false negative samples are distributed in the center of the true positive sample cluster, which means that CLML has effectively discover and narrow the distance between the missing labels and the true positive labels.

The above comparison results all prove the superiority of CLML in maximizing inter-class variance, minimizing intra-class variance and mining missing labels.

\section{Conclusion}
In this paper, we present a new contrastive loss for MLML tasks. With the help of a label correction mechanism to identify missing labels, CLML can accurately bring images close to their true positive images and false negative images, far away from their true negative images. Interestingly, low-rank global and local label dependencies are also preserved by the proposed contrastive loss in the latent representation space. Our empirical results showed that it can effectively increase the performance of other loss functions and achieve state-of-the-art classification performance on three benchmarks.
\ifCLASSOPTIONcaptionsoff
  \newpage
\fi


\end{document}